\documentclass[10pt,twocolumn,letterpaper]{article}

\usepackage{wacv}
\usepackage{times}
\usepackage{epsfig}
\usepackage{graphicx}
\usepackage{amsmath}
\usepackage{amssymb}

\usepackage[accsupp]{axessibility}

\usepackage{glossaries}
\usepackage{xcolor}
\newacronym{cnn}{CNN}{Convolutional Neural Network}
\newacronym{e2e}{e2e}{end-to-end}
\newacronym{gps}{GPS}{Global Positioning System}
\newacronym{gcd}{\emph{GCD}}{Great Circle Distance}
\newacronym{ah}{\emph{SemP}}{semantic partitioning}
\newacronym{osm}{OSM}{Open Street Map}
\newacronym{ci}{\emph{ci}}{Concept Influence}
\newacronym{sgd}{SGD}{Stochastic Gradient Descent}
\newacronym{sgdr}{SGDR}{Stochastic Gradient Descent with Warm Restarts}
\newacronym{mp16}{\emph{MP-16}}{Media\-Eval Placing Task 2016}
\newacronym{yfcc100m}{\emph{YFCC100M}}{Yahoo Flickr Creative Commons 100 Million}
\newacronym{yfcc25600}{\emph{YFCC-Val26k}}{Yahoo Flickr Creative Commons 25600 Validation}
\newacronym{mvmf}{MvMF}{\emph{Mixture of von-Mises Fisher}}
\newacronym{hrnet}{\emph{HRNetV2}}{high-resolution network}

\usepackage{tabularx}
\usepackage{booktabs}
\usepackage{multirow}
\usepackage[numbers,sort,compress]{natbib}

\def\wacvPaperID{293} 

\wacvfinalcopy 

\ifwacvfinal
\def\assignedStartPage{1} 
\fi

\ifwacvfinal
\usepackage[breaklinks=true,colorlinks,bookmarks=false]{hyperref} 
\else
\usepackage[pagebackref=true,breaklinks=true,colorlinks,bookmarks=false]{hyperref}
\fi

\ifwacvfinal
\else
\fi

\begin{document}

\title{Interpretable Semantic Photo Geolocation}

\author{Jonas Theiner$^1$ \qquad Eric Müller-Budack$^2$ \qquad Ralph Ewerth$^{1,2}$  \\\\
$^1$L3S Research Center, Leibniz University Hannover, Hannover, Germany \\ $^2$TIB -- Leibniz Information Centre for Science and Technology, Hannover, Germany \\
{\tt\small theiner@l3s.de} \qquad {\tt\small $\{$eric.mueller, ralph.ewerth$\}$@tib.eu}
}

\maketitle

\begin{abstract}
Planet-scale photo geolocalization is the complex task of estimating the location depicted in an image solely based on its visual content. Due to the success of convolutional neural networks~(CNNs), current approaches achieve super-human performance. However, previous work has exclusively focused on optimizing geolocalization accuracy. Due to the black-box property of deep learning systems, their predictions are difficult to validate for humans. State-of-the-art methods treat the task as a classification problem, where the choice of the classes, that is the partitioning of the world map, is crucial for the performance. In this paper, we present two contributions to improve the interpretability of a geolocalization model:~(1) We propose a novel semantic partitioning method which intuitively leads to an improved understanding of the predictions, while achieving state-of-the-art results for geolocational accuracy on benchmark test sets;~(2)~We introduce a metric to assess the importance of semantic visual concepts for a certain prediction to provide additional interpretable information, which allows for a large-scale analysis of already trained models. Source code and dataset are publicly available\footnote{\scriptsize \url{https://github.com/jtheiner/semantic_geo_partitioning}}.
\end{abstract}

\section{Introduction}

\begin{figure}[t!]
\begin{center}
  \includegraphics[width=1.0\linewidth]{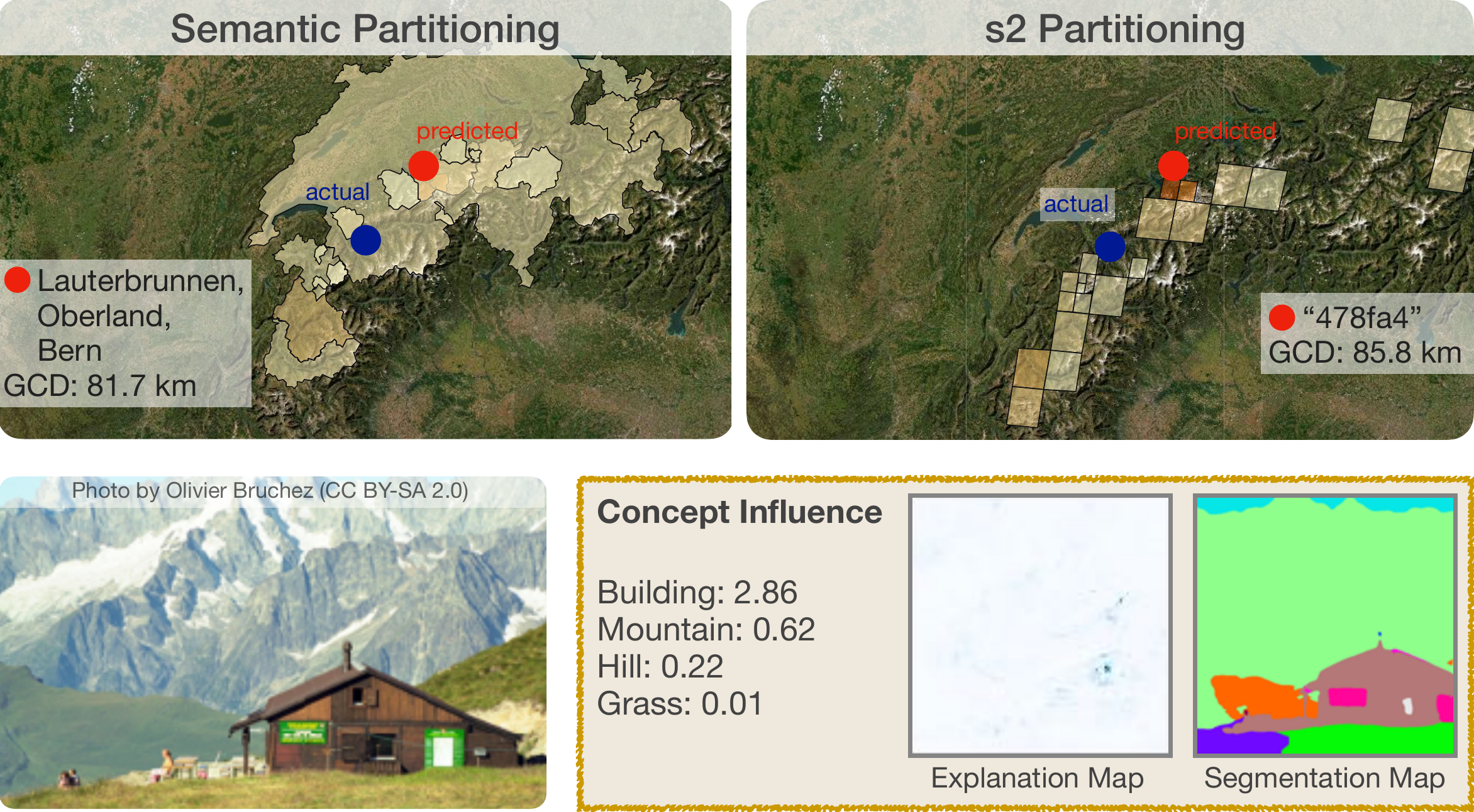}
\end{center}
  \caption{Example output of a geolocalization model with our proposed semantic partitioning for more explainable geolocation estimation~(left) compared to an existing partitioning approach~(right).
  Additionally, we measure the influence of visible concepts on the prediction of a model given an explanation and segmentation map.}
\label{fig:motivation}
\end{figure}

Image geolocalization is the challenging task of predicting the location of a photo in form of \acrshort{gps} coordinates based only on its visual content.
Almost all state-of-the-art approaches for planet-scale image geolocalization~\cite{Weyand2016PlanetPhotoGeolocationWithConvolutionalNeuralNetworks, MuellerBudack2018GeolocationEstimationOfPhotosUsingAHierarchicalModelAndSceneClassification, Seo2018CPlaNetEnhancingImageGeolocalizationByCombinatorialPartitioningOfMaps} define the task as a classification problem, where the earth is divided into geographical cells~(called \emph{partitioning}), 
and train \Glspl{cnn} with a huge amount of labeled data in an \acrlong{e2e} fashion. 
This strategy and the large amount of parameters in the networks turn them into a kind of black-box-systems, whose reasoning and predictions are not comprehensible~--
~making it necessary to develop methods to understand their decisions~\cite{Gilpin2018ExplainingExplanationsAnOverviewOfInterpretabilityOfMachineLearning, Guidotti2019ASurveyOfMethodsForExplainingBlackBoxModels}.
This is particularly a requirement for geolocalization systems for two reasons. First, humans are far worse at estimating locations than current deep learning approaches~\cite{Weyand2016PlanetPhotoGeolocationWithConvolutionalNeuralNetworks}; second, research has focused exclusively on maximizing localization accuracy, but lacks proposals for interpretable and explainable models. 

While many approaches~(e.g., \cite{Arandjelovic2018NetVLADCNNArchitectureForWeaklySupervisedPlaceRecognition, Sattler2016LargeScaleLocationRecognitionAndTheGeometricBurstinessProblem, Noh2017LargeScaleImageRetrievalWithAttentiveDeepLocalFeatures, Tzeng2013UserDrivenGeolocationOfUntaggedDesertImageryUsingDigitalElevationModels, Saurer2016ImageBasedGeoLocalizationInTheAlps}) restrict the problem of photo geolocalization to a part of the earth~(e.g., landmarks or mountains), predicting coordinates at planet-scale without any restrictions is more complex. 
Landmarks (usually tourist attractions) can partly be verified by humans, whereas many photos give little indication of the actual place or region. 
As a result, the question arises which features have been learned and which image features are relevant for a given prediction.
Furthermore, the quadratic boundaries of the \emph{s2} partitioning~\cite{Weyand2016PlanetPhotoGeolocationWithConvolutionalNeuralNetworks} are arbitrary~(see Figure~\ref{fig:motivation}) and the cells of \emph{CPlaNet}~\cite{Seo2018CPlaNetEnhancingImageGeolocalizationByCombinatorialPartitioningOfMaps} are initialized randomly which is counter-intuitive with regard to comprehensibility.
Following these considerations, a \acrshort{cnn}-based approach for geolocation estimation should therefore also be assessed with regard to the interpretability of its results. 

In this paper, we address this issue and introduce a novel \acrfull{ah} method where the cells are not rectangular or arbitrarily shaped as in previous approaches~\cite{Weyand2016PlanetPhotoGeolocationWithConvolutionalNeuralNetworks, Seo2018CPlaNetEnhancingImageGeolocalizationByCombinatorialPartitioningOfMaps, MuellerBudack2018GeolocationEstimationOfPhotosUsingAHierarchicalModelAndSceneClassification}. Instead, the partitioning considers real and interpretable locations derived from territorial~(e.g., streets, cities, or countries), natural~(e.g., rivers, mountains), or man-made boundaries~(e.g., roads, railways, or buildings) extracted from \Gls{osm}~\cite{OpenStreetMap2017PlanetDumpRetrievedFromHttpsplanetosmorg} data~(Figure~\ref{fig:motivation}). 
This partitioning better reflects location entities and we argue that photos taken within these boundaries also more likely share similar geographic attributes.
As a result, training and output of a model are more comprehensible to humans by default, while at the same time, state-of-the-art results on common test sets are achieved.
In addition, we suggest a \emph{concept influence} metric to investigate the post-hoc interpretability by measuring the influence of semantic visual concepts on individual predictions~(example in Figure~\ref{fig:motivation}). 
Experimental results show that the novel semantic partitioning method achieves (at least) state-of-the-art performance, while the concept influence score provides insights which visual concepts contribute to correct and incorrect~(or misleading) predictions. 

The rest of the paper is organized as follows. Related work for photo geolocation estimation is reviewed in Section~\ref{sec:rw}. The novel semantic partitioning method and \emph{concept influence} score are described in Section~\ref{sec:explainable_geolocation}, while experimental results are reported in Section~\ref{sec:experiments}, both for accuracy and interpretability of the results.
Section~\ref{sec:conclusion} summarizes the paper and outlines areas of future work.


\section{Related Work}\label{sec:rw}

Whereas only few approaches~\cite{Weyand2016PlanetPhotoGeolocationWithConvolutionalNeuralNetworks, Vo2017RevisitingIm2gpsInTheDeepLearningEra, MuellerBudack2018GeolocationEstimationOfPhotosUsingAHierarchicalModelAndSceneClassification, Seo2018CPlaNetEnhancingImageGeolocalizationByCombinatorialPartitioningOfMaps, Izbicki2019ExploitingTheEarthsSphericalGeometryToGeolocateImages} are applicable at planet-scale without limitations, the majority simplifies the task of geolocalization, for example, by predicting landmarks and cities~\cite{Arandjelovic2018NetVLADCNNArchitectureForWeaklySupervisedPlaceRecognition, Zheng2009TourTheWorldBuildingAWebScaleLandmarkRecognitionEngine, Sattler2016LargeScaleLocationRecognitionAndTheGeometricBurstinessProblem, Noh2017LargeScaleImageRetrievalWithAttentiveDeepLocalFeatures}, natural areas~\cite{Baatz2012LargeScaleVisualGeoLocalizationOfImagesInMountainousTerrain, Lin2013CrossViewImageGeolocalization, Schindler2007CityScaleLocationRecognition, Tzeng2013UserDrivenGeolocationOfUntaggedDesertImageryUsingDigitalElevationModels, Saurer2016ImageBasedGeoLocalizationInTheAlps}, or geo-related attributes~\cite{Lee2015PredictingGeoInformativeAttributesInLargeScaleImageCollectionsUsingConvolutionalNeuralNetworks}.

Previous work uses either image retrieval approaches or models the task as a classification problem.
The task of geolocation estimation at planet-scale has an overlap with methods from instance-level image retrieval~\cite{Noh2017LargeScaleImageRetrievalWithAttentiveDeepLocalFeatures, Arandjelovic2018NetVLADCNNArchitectureForWeaklySupervisedPlaceRecognition, Gordo2016DeepImageRetrievalLearningGlobalRepresentationsForImageSearch, Radenovic2016CNNImageRetrievalLearnsFromBoWUnsupervisedFineTuningWithHardExamples, Tolias2016ParticularObjectRetrievalWithIntegralMaxPoolingOfCNNActivations} where benchmark datasets consist of popular places,  landmarks, and tourist attractions~\cite{Philbin2007ObjectRetrievalWithLargeVocabulariesAndFastSpatialMatching, Philbin2008LostInQuantizationImprovingParticularObjectRetrievalInLargeScaleImageDatabases, Jegou2008HammingEmbeddingAndWeakGeometricConsistencyForLargeScaleImageSearch} which can be verified by humans.
Common to all is the usage of triplet ranking or contrastive embeddings to learn discriminative image representations, whereas \citet{Liu2019StochasticAttractionRepulsionEmbeddingForLargeScaleImageLocalization} introduce an alternative loss function. 
These representations are used to retrieve the most similar images in a reference database in order to determine the geolocation as proposed by \textit{Im2GPS}~\cite{Hays2008IM2GPSEstimatingGeographicInformationFromASingleImage, Hays2015LargeScaleImageGeolocalization}.
\citet{Weyand2016PlanetPhotoGeolocationWithConvolutionalNeuralNetworks} introduce the classification approach \emph{PlaNet}, where a \acrshort{gps} coordinate is mapped to a discrete class label using a quad-tree approach that divides the surface of the earth into distinct regions using the \emph{s2} geometry library~\cite{S2Library}. 
This \emph{s2} partitioning is used at multiple spatial scales to exploit hierarchical knowledge~\cite{Vo2017RevisitingIm2gpsInTheDeepLearningEra, MuellerBudack2018GeolocationEstimationOfPhotosUsingAHierarchicalModelAndSceneClassification}.
A pre-classification step assigns a photo to one of three scene types~(natural, urban, indoor) and leads to improvements~\cite{MuellerBudack2018GeolocationEstimationOfPhotosUsingAHierarchicalModelAndSceneClassification}. 
\citet{Seo2018CPlaNetEnhancingImageGeolocalizationByCombinatorialPartitioningOfMaps} propose a combinatorial partitioning where the overlaps of multiple coarse-grained partitionings create one fine-grained partitioning.
\citet{Izbicki2019ExploitingTheEarthsSphericalGeometryToGeolocateImages} introduce the \gls{mvmf} loss function for the classification layer that exploits the earth's spherical geometry and refines the geographical cell shapes in the partitioning. 
\citet{kordopatis2021leveraging} combine classification~\cite{Izbicki2019ExploitingTheEarthsSphericalGeometryToGeolocateImages, MuellerBudack2018GeolocationEstimationOfPhotosUsingAHierarchicalModelAndSceneClassification} and retrieval techniques to leverage the advantages of each approach, i.e., learning global knowledge from classification and exploit local features via retrieval~(landmark matching).


\section{Interpretable Semantic Photo Geolocation}
\label{sec:explainable_geolocation}
\begin{figure}[bt!]
\begin{center}
  \includegraphics[width=1.0\linewidth]{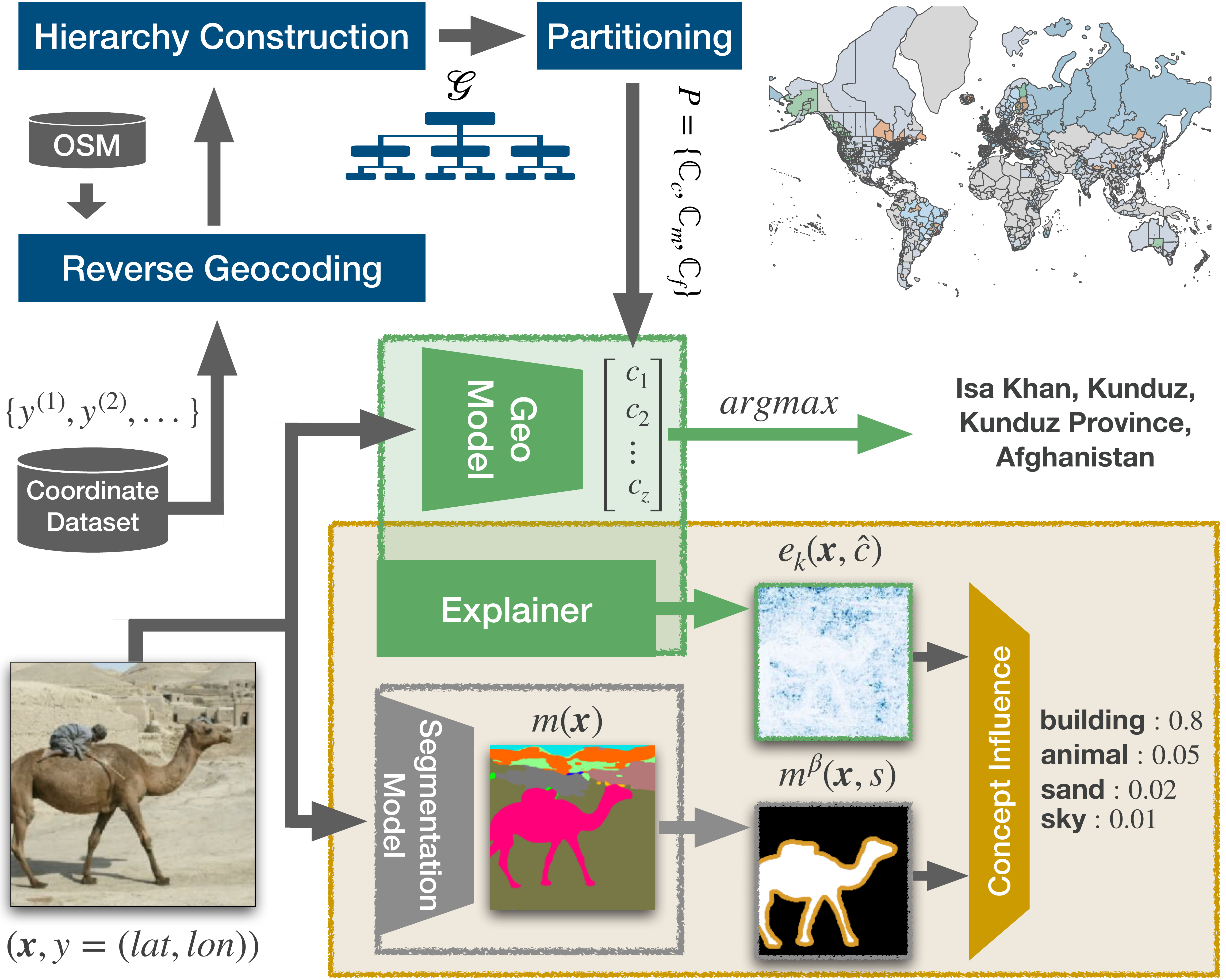}
\end{center}
  \caption{Overview of the bipartite system: Top (blue): Workflow to create a \acrfull{ah}, bottom (orange): Components to measure the \emph{concept influence} for individual samples.}
\label{fig:workflow}
\end{figure}

As the discussion of related work reveals, all approaches~(classification and retrieval) rely on features from \Glspl{cnn} that are learned with the use of a \emph{partitioning}. Their predictions are difficult to interpret and the construction of the \emph{partitioning} is crucial for the system performance~\cite{Vo2017RevisitingIm2gpsInTheDeepLearningEra,Seo2018CPlaNetEnhancingImageGeolocalizationByCombinatorialPartitioningOfMaps}.
The following two subsections address two issues with regard to interpretability.
First, a novel partitioning method is proposed that relies on data that are derived from a geographic database where metadata about many regions and places such as their size or exact boundaries is provided.
Second, a method is presented to automatically assess image features that are relevant for a model's decision based on semantic visual concepts like \emph{waterfall} or \emph{person}. 
Their workflows and connections are outlined~in~Figure~\ref{fig:workflow}.

\subsection{Semantic Partitioning}
\label{sec:admin_part}

State-of-the-art methods for photo geolocalization rely on classification approaches~\cite{Seo2018CPlaNetEnhancingImageGeolocalizationByCombinatorialPartitioningOfMaps, MuellerBudack2018GeolocationEstimationOfPhotosUsingAHierarchicalModelAndSceneClassification}, where the design of the classification layer is crucial for the model's output with respect to prediction accuracy, but also regarding the information that is provided to users. 
The main idea is to divide the earth surface into a discrete set of classes~$\mathbb{C}$ based on the dataset distribution to then train a classification network~\cite{Weyand2016PlanetPhotoGeolocationWithConvolutionalNeuralNetworks}. 
We follow the same idea, but after all our cells cover territorial borders~(e.g., countries, cities), natural geological boundaries~(e.g., rivers, mountains) or man-made barriers~(e.g., roads or railways that separate districts).
In addition to an improved understanding of the created cells, the assumption is that a \acrshort{cnn} learns better image representations, since the resulting geographic cells might better represent locations and are thus more distinguishable.
The following steps describe formally how that \acrfull{ah} is constructed.

\subsubsection{Reverse Geocoding \& Hierachy Construction}
\label{subsec:reverse_geocoding}

Following the idea of classification, a mapping is needed from the continuous \acrshort{gps} coordinate space to a discrete set of existing locations which is called reverse geocoding. 
Frameworks for reverse geocoding generate an address vector, e.g., (\emph{Long Beach, Los Angeles County, California, USA})
with a coordinate as input. We choose \textit{Nominatim}~\cite{nominatim} since it is open source software and relies on \acrshort{osm}. 
Formally, a reverse geocoder maps each coordinate $y^{(i)} = (\text{latitude}, \text{longitude})$ in a dataset $\mathbb{D}=\{ y^{(1)}, y^{(2)}, \hdots\}$ to an address vector $\boldsymbol{l}^{(i)}=(l_1, \hdots, l_u)$ of arbitrary length $u$ and is ordered from fine to coarse. 
This mapping is denoted as $\mathbb{D}^l=\{\boldsymbol{l}^{(1)}, \boldsymbol{l}^{(2)}, \hdots\}$.

\textbf{Hierarchy Construction:}
Since hierarchical knowledge is valuable with respect to performance~\cite{Weyand2016PlanetPhotoGeolocationWithConvolutionalNeuralNetworks, MuellerBudack2018GeolocationEstimationOfPhotosUsingAHierarchicalModelAndSceneClassification} and all necessary information is already provided by the reverse geocoder, we construct a hierarchy similar to the \emph{s2} library~\cite{Weyand2016PlanetPhotoGeolocationWithConvolutionalNeuralNetworks} but with semantically meaningful nodes and edges.
In order to create a partitioning from the obtained addresses $\mathbb{D}^l$, it is required to build a hierarchy where each discrete location~$l$~(e.g., \emph{Long Beach}) can be assigned to its next coarser distinct location~(e.g., \emph{Los Angeles County}). 
A directed (multi-)~graph $\mathcal{G}=(V, E)$ can be constructed using all edges that occur in the mapping $\mathbb{D}^l$.
The total number of nodes corresponds to the number of locations; an edge exists between two adjacent nodes encoded in the mapping $\mathbb{D}^l$, i.e., $(l_i, l_{i+1}) \forall i \in [1, \hdots, u-1]$ for every $\boldsymbol{l}=(l_1, \hdots, l_u) \in \mathbb{D}^l$. Nodes without outgoing edges~(roots) usually correspond to countries. 
Ideally, $\mathcal{G}$ consists only of trees, where exactly one parent node is assigned to each node with the exception of the root nodes. 
Otherwise, $\mathcal{G}$ must be transformed into a hierarchy. 
For each location only the most frequent outgoing edge is kept. Reasons for multiple parents are possible incorrect assignments for some instances or missing assignments that cause shortcuts. 
Therefore, the mapping $\mathbb{D}^l$ is subsequently replaced with the locations from the shortest path of the finest location $l_1$ in $(l_1, ..., l_u)=\boldsymbol{l} \quad \forall \boldsymbol{l} \in \mathbb{D}^l$ to its root node in the hierarchy $\mathcal{G}$ and is referred to as $\mathbb{D}^{l^{*}}$.

\subsubsection{Partitioning Construction \& Cell Assignment}
\label{subsec:partitioning_construction}
In order to create the \acrshort{ah}, the set of coordinates~$\mathbb{D}$ is first transformed to a~(hierarchical) multi-label dataset as described in Section~\ref{subsec:reverse_geocoding}.
A valid partitioning would be to consider only the finest location $l^{(i)}_1$ for each $\boldsymbol{l}^{(i)}$.
In practice, a huge number of classes is not manageable and previous work (e.g.,~\cite{Weyand2016PlanetPhotoGeolocationWithConvolutionalNeuralNetworks}) controls the granularity of a partitioning. 
This choice of granularity entails a trade-off problem. 
While fewer but larger~(in terms of geographic area) cells decrease the geospatial resolution of the model outputs, more but smaller cells are more challenging to distinguish. 
They also make the model susceptible to overfitting due to the lower number of available training images per cell~\cite{Seo2018CPlaNetEnhancingImageGeolocalizationByCombinatorialPartitioningOfMaps}. 
Moreover, geographic information at different spatial resolutions are important to identify locations of varying granularity~(e.g., buildings, cities, or countries).
To construct a partitioning $\mathbb{C}$ at a certain spatial level, we first delete all locations from the derived hierarchy $\mathcal{G}$ with less than $\tau_{\text{min}}$ images. As a result, we derive a mapping $\mathbb{D}^{l^{*}}$ with the remaining locations in the graph. 
%
The finest locations in $\mathbb{D}^{l^{*}}$ form a partitioning, i.e., all $l_1$ from all $\boldsymbol{l} \in \mathbb{D}^{l^{*}}$. 
To assign a dataset $\mathbb{D}_{\text{new}}$ to classes from a created partitioning $\mathbb{C}$, two steps are necessary. 
First, the same reverse geocoder has to create an initial assignment $\mathbb{D}^{l^{*}}_{\text{new}}$ and these discrete locations are filtered by the available locations (now classes) from the partitioning $\mathbb{C}$. Given the $i$-th sample $\boldsymbol{l}^{(i)} \in \mathbb{D}^{l^{*}}_{\text{new}}$, the location $l^{(i)}_1$ from $\boldsymbol{l}^{(i)}$ corresponds to the finest available one according to the partitioning $\mathbb{C}$.
%

\subsubsection{Learning \& Inference}
With the classes~$\mathbb{C}$ obtained from the presented partitioning method, a \acrshort{cnn} can be trained directly on the classification task using the cross-entropy loss ($\mathcal{L_\text{CE}}$) where the number of classes corresponds to the number of cells of the partitioning. 
Initially, only a dataset of image-coordinate pairs is necessary where the coordinates are transformed to classes according to Section~\ref{subsec:partitioning_construction}.
Multiple partitionings can be combined to force the model to learn some kind of hierarchical knowledge. 
Given a tuple of partitionings $\mathcal{P}=(\mathbb{C}_1, ..., \mathbb{C}_n)$ which differ only in $\tau_{\text{min}}$~(i.e., controls the number of classes) and are ordered from fine to coarse, each cell in $\mathbb{C}_i$ can be assigned to its corresponding cell in $\mathbb{C}_{i+1}$ by exploiting the hierarchy $\mathcal{G}$. One fully-connected layer per partitioning $\mathbb{C}_i$ is added on top of an appropriate \acrshort{cnn} architecture. During training, the multi-partitioning classification loss is defined as the sum of all individual losses per partitioning $\mathcal{L}_\text{CE}^\text{multi}=\sum_{i=1}^{|\mathcal{P}|}\mathcal{L}_{\text{CE}}^i$.
During inference, the class at the finest partitioning with the highest probability after applying the softmax function corresponds to the predicted cell~$\hat{c}$.
%
%
We use the average \acrshort{gps} coordinate of the assigned samples from~$\mathbb{D}$ during the partitioning process from the respective class as geolocation prediction.

\subsection{Measuring the Input Feature Importance}
\label{sec:ci}
In the task of photo geolocalization, we do not know which image regions are crucial for the model's prediction and cannot validate the decisions. While methods for the visualization of feature attributions have been researched in recent years, the main focus was on object recognition where the highlighted areas are comprehensible at least to humans~\cite{Ancona2017AUnifiedViewOfGradientBasedAttributionMethodsForDeepNeuralNetworks}. Inspired by these approaches, we propose a method to measure the influence of specific objects (e.g., \emph{vehicle} or \emph{person}) and semantic image regions (e.g., \emph{sky} or \emph{ground}) regarding the model's prediction. 
The goal is not to identify a concrete concept that is responsible for the prediction -- which would be counter intuitive, since a decision should not be reduced to image regions exclusively. Rather it can be helpful to estimate the overall impact of a given semantic concept, to identify misleading concepts, or to provide explanatory information in form of a more comprehensible~(text and quantitative values) and summarized~(reduced to relevant concepts) explanation map to users.
Attribution maps also provide an importance value per pixel, but also a lot of noise~\cite{Smilkov2017SmoothGradRemovingNoiseByAddingNoise} and allow the observer freedom in the interpretation.

\paragraph{Required Components:}
Formally, an input image $\boldsymbol{x} \in \mathbb{R}^{w \times h \times d}$ and a \acrshort{cnn}
~$\Psi$ are required.
Only two components are needed to calculate the influence of concepts on the prediction.
An \emph{explanation map} $e$ assigns an importance score to each input pixel of $\boldsymbol{x}$ for a certain prediction~\cite{Sundararajan2017AxiomaticAttributionForDeepNetworks, Simonyan2014DeepInsideConvolutionalNetworksVisualisingImageClassificationModelsAndSaliencyMaps, Springenberg2015StrivingForSimplicityTheAllConvolutionalNet}, e.g., a target class $\hat{c}$ in case of a classification model. 
The maximum over the~(color) channel dimension $d$ is taken as only image regions are of interest, hence we define $e: \mathbb{R}^{w \times h \times d} \mapsto \mathbb{R}^{w \times h}$.
Please note, that usually only the gradients of the model $\Psi$ have to be accessible for the calculation.
A \emph{segmentation map}~$m$ divides image areas into semantic groups~(e.g., a region, object, or texture).
The segmentation mask for one concept~$s \in \mathbb{S}$ is the indicator function where the presence of $s$ on a pixel is indicated by $m: \mathbb{R}^{w \times h \times d} \mapsto \{0, 1\}^{w \times h}$ and denoted as $m(\boldsymbol x, s)$.

Assuming that the area of the segmentation boundaries, i.e., the border between two concepts is of interest for the geo model resulting in activations in the explanation map, the active area of the binary mask $m(\boldsymbol{x}, s)$ for concept $s$ can be enlarged $\beta$ pixels around its shape boundary using a morphological dilation,  and is denoted as $m^\beta(\boldsymbol{x}, s)$ as seen in Figure~\ref{fig:workflow}~(orange colorized area around the camel's surface).

\paragraph{Concept Influence:}
The aim is to measure the influence of a specific concept $s$ using the explanation map $e(\boldsymbol{x}, \hat{c})$ and the segmentation map $m(\boldsymbol{x}, s)$ for a specific concept $s$. As stated by~\citet{Ghorbani2019InterpretationOfNeuralNetworksIsFragile}, in many settings only the most important features are of explanatory interest. They compute the pixel-wise intersection of the $k$ most important features from $e(\boldsymbol{x}, \hat{c})$ to measure the difference between two explanation maps (top-$k$ intersection). However, this is done for a slightly different purpose, that is generating and evaluating manipulated explanation maps. 
Inspired by this, we adapt this measure to define the influence of a concept $s$ visible in image~$\boldsymbol{x}$ with respect to a geoprediction $\hat{c}$ of the model. We define the pixel-wise intersection $tki$ between the binary segmentation mask $m(\boldsymbol{x}, s)$ and the binary mask of the top-$k$ features $e_k(\boldsymbol{x}, \hat{c})$ as
\begin{equation}
\label{eq:tki}
    tki = \frac{1}{k} \sum_{i=1}^w \sum_{j=1}^h m(\boldsymbol{x}, s)_{i,j} \wedge e_k(\boldsymbol{x}, \hat c)_{i,j}
\end{equation}
where $m(\boldsymbol{x}, s)$ and $e_k(\boldsymbol{x}, \hat{c})$ are both in $\{0,1\}^{w \times h}$ and~$\wedge$~is the pixel-wise boolean \emph{and} operation.
For instance, if all top-$k$ pixels are within the shape of the concept $s$ then $tki=1$. 
In our experiments, we set the parameter $k$ to 1,000 as proposed by~\citet{Ghorbani2019InterpretationOfNeuralNetworksIsFragile}.
As large objects or areas are preferred, a normalization step is crucial for application. The defined top-$k$ intersection (\emph{tki}) is hence
normalized by the relative size of the concept which is defined as:
\begin{equation}
\overline{s} := \frac{1}{wh} \sum_{i=1}^w \sum_{j=1}^h \big( m(\boldsymbol{x},s) \big)_{i,j}
\end{equation}
The resulting score is the definition of the \emph{concept influence} (\emph{ci}) metric~$ci \big( m(\boldsymbol{x}, s), e_k(\boldsymbol{x}, \hat{c}) \big) = tki / \overline{s}$.
A \emph{ci} score of less than or equal to one means that the top-$k$ pixels of the explanation map are more likely to be in other regions of the image, i.e., class $s$ has little or no influence on the final prediction. The \emph{ci} score indicates whether a concept $s$ contains relatively large number of activations of the explainer $e$.
When fixing the minimum required relative concept size to $0 < \overline{s}_{\text{min}} < 1$, \emph{ci} $\in [0, \frac{1}{\overline{s}_{\text{min}}}]$ is well defined and only those concepts are considered for the calculation that cover at least this area.
Additionally, we assume that small concepts that cover only a minimal area in the image to be irrelevant or noisy and set $\overline{s}_{\text{min}}=0.05$ in our experiments.

Finally, given a segmentation map $m$ and an explanation map $e$ for model $\Psi$, the introduced metric \emph{ci} automatically measures the impact of semantic image regions for the prediction.

\section{Experimental Results}
\label{sec:experiments}
In this section, the proposed partitioning method is evaluated with respect to geolocational accuracy and its capability of providing an improved interpretability (Section~\ref{exp:ah}). Afterwards, the \emph{concept influence} metric introduced in Section~\ref{sec:ci} is evaluated (Section~\ref{exp:ci}). 

\subsection{Semantic Partitioning}
\label{exp:ah}
We demonstrate the capability of our approach through a comparison with several state-of-the-art models, including a model that also exploits the hierarchical knowledge from multiple partitionings~\cite{MuellerBudack2018GeolocationEstimationOfPhotosUsingAHierarchicalModelAndSceneClassification}, on three benchmark datasets. 

\subsubsection{Experimental Setup}
\textbf{Datasets \& Evaluation Metric:}
We utilize the \acrfull{mp16}~\cite{Larson2017TheBenchmarkingInitiativeForMultimediaEvaluationMediaEval2016} dataset which is a subset from the \acrfull{yfcc100m}~\cite{Thomee2016YFCC100MTheNewDataInMultimediaResearch} both for partitioning construction and training. Its only restriction is that an image contains a \acrshort{gps} coordinate, thus it contains images of landmarks, landscape images, but also images with little to no geographical cues. Like~\citet{Vo2017RevisitingIm2gpsInTheDeepLearningEra}, images are excluded from training if there are photos taken by the same authors in the \emph{Im2GPS3k} test set and duplicates are removed, resulting in a dataset size of 4,723,695 image-coordinate pairs. 
For validation, a randomly sampled subset of 25,600 images from \acrshort{yfcc100m} without overlap to the training images is created and denoted as \acrshort{yfcc25600}.
For testing, we focus on three popular benchmark datasets: 
\emph{YFCC4K}~\cite{Vo2017RevisitingIm2gpsInTheDeepLearningEra} comes from the same image domain as the training dataset but is designed for general computer vision tasks making the test set more challenging. 
In contrast, the \emph{Im2GPS}~\cite{Hays2008IM2GPSEstimatingGeographicInformationFromASingleImage} and \emph{Im2GPS3k}~\cite{Vo2017RevisitingIm2gpsInTheDeepLearningEra} datasets contain some landmarks, but the majority of images is recognizable only in a generic sense like landscapes.

For evaluation, the geolocational accuracy at multiple error levels, i.e., the tolerable error in terms of distance from the predicted~$l_{\text{pred}}$ to the ground-truth location~$l_{\text{gt}}$ is calculated~\cite{Weyand2016PlanetPhotoGeolocationWithConvolutionalNeuralNetworks, Vo2017RevisitingIm2gpsInTheDeepLearningEra}.
Formally, the geolocational accuracy $a_r$ at scale $r$ (in km) is defined as follows for a set of $N$ samples:
\begin{equation}
    a_r \equiv \frac{1}{N} \sum_{i=1}^{N} u\left(d(l_{\text{gt}}^{(i)}, l_{\text{pred}}^{(i)}) < r \right),
\end{equation}
where the distance function is the \acrfull{gcd} and $u(\cdot)$ is the indicator function whether the distance is smaller than the tolerated radius $r$. 

\textbf{Partitioning Parameters:}
First, the coordinates from the \acrshort{mp16} are transformed to a multi-label dataset containing 2,191,616 unique locations. To initially reduce the number, we delete all locations with less then 50 images.
resulting in manageable 46,240 unique locations.
Due to the proven importance of a multi-partitioning~\cite{Vo2017RevisitingIm2gpsInTheDeepLearningEra, Seo2018CPlaNetEnhancingImageGeolocalizationByCombinatorialPartitioningOfMaps, MuellerBudack2018GeolocationEstimationOfPhotosUsingAHierarchicalModelAndSceneClassification}, we directly evaluate this setting. 
For a fair comparison, we construct a multi-partitioning that consists of three individual partitionings~(coarse, middle, fine) with a similar total number of unique classes compared to \citet{MuellerBudack2018GeolocationEstimationOfPhotosUsingAHierarchicalModelAndSceneClassification} and follow their notation.
To construct a multi-partitioning, several thresholds~$\tau_{\text{min}}$ can be applied to get a similar number of classes, as shown in Table~\ref{tab:ah_paramter_setting}.
For this reason, we select the model that performs best on the validation set for the comparison and evaluation on the test sets. 
Furthermore, we investigate three additional settings:
(1) To keep the parameters fixed, but applying one filter, i.e., utilization of locations that are associated with geographic area stored as a (multi-)polygon according to \acrshort{osm}~(denoted as $\acrshort{ah}_a$);
(2) testing the hierarchical prediction variant ($f$ vs. $f^*$); and 
(3) testing the scalability by doubling the number of classes.

\begin{table}[b!]
\caption{Experimental results on the validation set of \acrshort{yfcc25600} for several multi-partitonings where $|\mathbb{C}|$ is the total number of unique classes.}
\label{tab:ah_paramter_setting}
\setlength{\tabcolsep}{2pt}
\begin{center}
\small
\fontsize{7}{10}\selectfont
\begin{tabularx}{\linewidth}{X|c|ccccc}
\toprule
\multirow{2}{*}{Configuration}  & \multirow{2}{*}{$|\mathbb{C}|$} & \multicolumn{5}{c}{$a_r$ [\%] @ km} \\ 
 &   & 1 & 25 & 200 & 750 & 2500 \\ 
\midrule

$\text{\acrshort{ah}}(\{100,125,150\}, f^*)$ & 14877 & 4.8 & 11.0 & 18.5 & 33.6 &  53.9 \\
$\text{\acrshort{ah}}(\{100,125,150\}, f)$ &  14877 & 7.5  & 15.8 & 23.8 & 38.0 & 56.6 \\ 

$\text{\acrshort{ah}}_a(\{100,150,250\}, f)$ & 12886 & 6.2 & 16.1 &  24.4  &  38.0  &  55.3 \\ 
$\text{\acrshort{ah}}(\{100,150,250\}, f)$ &  15127 & 6.6 & 16.4 & 24.0 & 37.6 &  55.4 \\ 
$\text{\acrshort{ah}}(\{100,125,250\}, f)$ &  15016 & 7.5 &  15.9  &  24.1  &  38.3  & 56.6 \\ 
$\text{\acrshort{ah}}_a(\{75,100,150\}, f)$ &  16808 &  6.6  &  16.4  & 24.0 & 37.6 &   55.4 \\ 

$\text{\acrshort{ah}}(\{50,75,100\}, f)$  & 34049 & 8.9  &  16.6  &  24.1  & 37.9 &  56.3 \\ \hline
$s_2(M, f^*)$ & 15606 & 6.8 & 16.4 & 24.6 & 38.4 &  56.8 \\
\bottomrule
\end{tabularx}
\end{center}
\end{table}

\textbf{Network Training \& Inference:}
We choose the commonly applied \emph{ResNet-50}~\cite{He2016DeepResidualLearningForImageRecognition, He2016IdentityMappingsInDeepResidualNetworks} and \emph{EfficientNet-B4}~\cite{tan2019efficientnet} as network architectures with an input dimension of $224\times224\times 3$ and $300\times300\times 3$, respectively.
As the \emph{ResNet-50} provides a good trade-off in terms of training time and performance, it is applied for the ablation study~(testing several partitioning parameters). 
The classification layers are added on top of the global pooling layer.
Instead of initializing the parameters of all models with \textit{ImageNet} weights, the weights from a model trained for ten epochs on countries is taken to derive features related to the problem. 
The \acrshort{sgd} method with an initial learning rate of $0.01$, a momentum of $0.9$, and weight decay of $10^{-4}$ is used to optimize for 15 epochs.
The learning rate is exponentially decreased by a factor of 0.5, initially after every three epochs, and every epoch from epoch 12 on. Training is performed with a batch size of 200 and validation is done after 512,000 images. 
Details for pre-training and image augmentation methods during training are reported in the appendix.
The model with the lowest loss $\mathcal{L}_{\text{CE}}^{\text{multi}}$ on the validation set is chosen. 
During inference, five crops are made and the mean prediction after applying softmax is taken.

\subsubsection{Results on the Validation Set}
The geolocational accuracies on the \acrshort{yfcc25600} validation set are reported in Table~\ref{tab:ah_paramter_setting}.
Results demonstrate that the exact choice of partitioning hyperparameters is not essential. All configurations with similar number of classes perform similarly well.
Surprisingly, the hierarchical prediction~($f^*$)~\cite{MuellerBudack2018GeolocationEstimationOfPhotosUsingAHierarchicalModelAndSceneClassification} is, in contrast to the assumptions, worse than considering only the finest partitioning~($f$). One technical reason might be the fundamental different underlying structure of the hierarchy $\mathcal{G}$ in contrast to the quad-tree~\cite{Weyand2016PlanetPhotoGeolocationWithConvolutionalNeuralNetworks}, resulting in a significantly lower depth and more variable number of child nodes. 
Humans may perceive locations hierarchically, but these coarse regions are not the ones with visually discriminative features.

\subsubsection{Benchmark Results}
From the models evaluated on the validation set, we select the one that has the best geolocational accuracy~($\text{\acrshort{ah}}(\{100,125,250\}, f)$), particularly for the error levels of 1\,km, 750\,km, and 2,500\,km.
Further, we assess the performance of one model considering only locations where geographic areas are available ($\text{\acrshort{ah}}_a(\{100,150,250\}, f$), and where the number of classes is doubled ($\text{\acrshort{ah}}(\{50,75,100\}, f)$).
As stated in the experimental setup, we test these configurations with two different \acrshort{cnn} backbones.
Quantitative results are reported for three test sets in Table~\ref{tab:admin_test_all}.

\begin{table}[bt!]
\caption{Geolocational accuracy ($a_r$) of \textbf{\acrshort{ah}} compared to several geolocalization approaches on common benchmark datasets. \emph{ResNet-50} and \emph{EfficientNet-B4} are applied for fair comparisons to the state of the art. 
Retrieval extensions and ensembles typically improve the performance and are colored gray.
}
\label{tab:admin_test_all}
\setlength{\tabcolsep}{4pt}
\setlength\extrarowheight{-1.4pt}
\begin{center}
\small
\fontsize{7}{10}\selectfont
\begin{tabularx}{\linewidth}{X|c|c|c|c|c}
\toprule

                                     &       \multicolumn{5}{c}{$a_r$ [\%] @ km} \\
 Approach            & 1  & 25    & 200    & 750    & 2500   \\
\midrule \midrule

\multicolumn{6}{c}{Im2GPS3k~\cite{Vo2017RevisitingIm2gpsInTheDeepLearningEra}~(2,997 images): geo-recognizable (generic)}  \\ \midrule
\text{[L]}7011C~\cite{Vo2017RevisitingIm2gpsInTheDeepLearningEra} & 4.0 & 14.8 & 21.4 & 32.6 & 52.4 \\
\text{[L]}kNN, $\sigma=4$~\cite{Vo2017RevisitingIm2gpsInTheDeepLearningEra}  & 7.2     & 19.4    & 26.9     & 38.9     & 55.9     \\
PlaNet~\cite{Weyand2016PlanetPhotoGeolocationWithConvolutionalNeuralNetworks} (rep.)~\cite{Seo2018CPlaNetEnhancingImageGeolocalizationByCombinatorialPartitioningOfMaps} & 8.5    & 24.8    & 34.3     & 48.4     & 64.6     \\
CPlaNet[1-5, PlaNet]~\cite{Seo2018CPlaNetEnhancingImageGeolocalizationByCombinatorialPartitioningOfMaps}             & 10.2    & 26.5    & 34.6     & 48.6     & 64.6     \\ 
MvMF$_{\text{B4}}$~(rep. \cite{kordopatis2021leveraging})                                                                          & 13.1 &  29.8    & 38.0    & 52.3   & 67.6     \\
$ISN(M, f^*, S_3)$~\cite{MuellerBudack2018GeolocationEstimationOfPhotosUsingAHierarchicalModelAndSceneClassification}   & \textcolor{gray}{10.5} &  \textcolor{gray}{28.0}    & \textcolor{gray}{36.6}     & \textcolor{gray}{49.7}     & \textcolor{gray}{66.0}     \\ 
$s2_{\text{B4}}(M, f^*)$~(rep.~\cite{kordopatis2021leveraging})~+~RRM   & \textcolor{gray}{13.2} &  \textcolor{gray}{29.1}    & \textcolor{gray}{37.8}     & \textcolor{gray}{52.0}     & \textcolor{gray}{68.1}     \\ 
MvMF$_{\text{B4}}$~(rep. \cite{kordopatis2021leveraging}) + RRM   & \textcolor{gray}{15.0} &  \textcolor{gray}{30.0}    & \textcolor{gray}{38.0}     & \textcolor{gray}{52.3}     & \textcolor{gray}{67.6}     \\ \hline
$s2_{\text{B4}}(M, f^*)$~(rep.)   & 11.5 & 30.8  &  41.0   & 55.7  &  70.8  \\ 
$\text{\textbf{\acrshort{ah}}}_{\text{B4}}(\{100,125,250\}, f)$                                                                                                 &  12.5 & \textbf{31.4} & \textbf{42.7} & \textbf{57.3}  &  \textbf{72.0} \\
$\text{\textbf{\acrshort{ah}}}_{\text{B4}}(\{50,75,100\}, f)$                                                                                                    &  \textbf{13.5} & 30.8 & 41.2 & 54.7  &  70.2 \\ \hline
$s2(M, f^*)$~\cite{MuellerBudack2018GeolocationEstimationOfPhotosUsingAHierarchicalModelAndSceneClassification}   & 9.7 & 27.0    & 35.6     & 49.2     & 66.0     \\
$s2(M, f^*)$~(rep.) &  10.0 & 27.0 & 36.5 &  50.9 &  67.2 \\ 
$\text{\textbf{\acrshort{ah}}}(\{100,125,250\}, f)$ & 11.1 & 27.1 & 36.7 & 50.4 &  66.1 \\
$\text{\textbf{\acrshort{ah}}}_a(\{100,150,250\}, f)$ & 9.6 & 26.9 & 36.8 & 49.7 &  65.1 \\
$\text{\textbf{\acrshort{ah}}}(\{50,75,100\}, f)$ &  11.5 & 27.0 & 36.3 & 49.3 &  65.9 \\ 
\midrule

\multicolumn{6}{c}{YFCC4k~\cite{Vo2017RevisitingIm2gpsInTheDeepLearningEra}~(4,536 images): no image restrictions}  \\ \midrule
\text{[L]}kNN, $\sigma=4$~\cite{Vo2017RevisitingIm2gpsInTheDeepLearningEra}  & 2.3     & 5.7    & 11.0     & 23.5     & 42.0     \\ 
PlaNet~\cite{Weyand2016PlanetPhotoGeolocationWithConvolutionalNeuralNetworks} (rep.)~\cite{Seo2018CPlaNetEnhancingImageGeolocalizationByCombinatorialPartitioningOfMaps} & 5.6    & 14.3    & 22.2     & 36.4     &  55.8     \\
CPlaNet[1-5, PlaNet]~\cite{Seo2018CPlaNetEnhancingImageGeolocalizationByCombinatorialPartitioningOfMaps}             & 7.9    & 14.8    & 21.9     & 36.4     & 55.5     \\ 
MvMF$_{\text{B4}}$~(rep. \cite{kordopatis2021leveraging})                                                                          & 6.8 &  14.4    & 21.9    & 37.5   & 56.4     \\ 
$s2_{\text{B4}}(M, f^*)$~(rep.~\cite{kordopatis2021leveraging})~+~RRM   & \textcolor{gray}{7.2} &  \textcolor{gray}{13.3}    & \textcolor{gray}{21.6}     & \textcolor{gray}{36.5}     & \textcolor{gray}{55.4}     \\ 
MvMF$_{\text{B4}}$~(rep. \cite{kordopatis2021leveraging}) + RRM   & \textcolor{gray}{7.9} &  \textcolor{gray}{14.3}    & \textcolor{gray}{21.9}     & \textcolor{gray}{37.4}     & \textcolor{gray}{56.5}     \\ \hline
$s2_{\text{B4}}(M, f^*)$~(rep.)  & 7.5  & 19.2  & 28.2  & 42.0  & 59.2   \\
$\text{\textbf{\acrshort{ah}}}_{\text{B4}}(\{100,125,250\}, f)$                                                                                                    &  9.4 & 20.3 & \textbf{30.6} & \textbf{44.8}  &  \textbf{61.2} \\ 
$\text{\textbf{\acrshort{ah}}}_{\text{B4}}(\{50,75,100\}, f)$                                                                                                    &  \textbf{12.1} & \textbf{22.3} & \textbf{30.6} & 43.5  &  60.4 \\ \hline

$s2(M, f^*)$~(rep.) &  6.6 & 16.4 &  24.1 & 36.8 & 55.1 \\ 
$\text{\textbf{\acrshort{ah}}}(\{100,125,250\}, f)$ & 7.3 & 15.3 & 23.9 &  37.2 &  54.3 \\
$\text{\textbf{\acrshort{ah}}}_a(\{100,150,250\}, f)$ & 6.1 & 15.8 & 23.9 & 36.8 &  52.6 \\
$\text{\textbf{\acrshort{ah}}}(\{50,75,100\}, f)$ &  9.3 &  17.1 &  24.1 & 36.9 &  54.3 \\ 
\midrule

\multicolumn{6}{c}{Im2GPS~\cite{Hays2008IM2GPSEstimatingGeographicInformationFromASingleImage} (237 images): majority shows landmarks}  \\ \midrule
Human~\cite{Vo2017RevisitingIm2gpsInTheDeepLearningEra}                                                         & \text{-}  & \text{-} & 3.8  & 13.9 & 39.3 \\
Im2GPS~\cite{Hays2008IM2GPSEstimatingGeographicInformationFromASingleImage}                               & \text{-}  & 12.0     & 15.0 & 23.0 & 47.0 \\
\text{[L]}kNN, $\sigma=4$, 28M~\cite{Vo2017RevisitingIm2gpsInTheDeepLearningEra}  & 14.4      & 33.3 & 47.7 & 61.6 & 73.4 \\
\text{[L]}7011C~\cite{Vo2017RevisitingIm2gpsInTheDeepLearningEra}                    & 6.8       & 21.9     & 34.6 & 49.4 & 63.7 \\ 
PlaNet~\cite{Weyand2016PlanetPhotoGeolocationWithConvolutionalNeuralNetworks}                                                & 8.4       & 24.5     & 37.6 & 53.6 & 71.3 \\
CPlaNet[1-5, PlaNet]~\cite{Seo2018CPlaNetEnhancingImageGeolocalizationByCombinatorialPartitioningOfMaps}                              & 16.5      & 37.1     & 46.4 & 62.0 & 78.5 \\

MvMF ($c=2^{17}$)~\cite{Izbicki2019ExploitingTheEarthsSphericalGeometryToGeolocateImages}  & 8.4       & 32.6     & 39.4 & 57.2 & 80.2 \\  
MvMF$_{\text{B4}}$~(rep. \cite{kordopatis2021leveraging})                                                                          & \textbf{19.8} &  \textbf{44.7}    & 55.7    & 67.5   & 81.9     \\ 
ISN(M, f*, $S_3$)~\cite{MuellerBudack2018GeolocationEstimationOfPhotosUsingAHierarchicalModelAndSceneClassification}  &  \textcolor{gray}{16.9}      &  \textcolor{gray}{43.0}     &  \textcolor{gray}{51.9} & \textcolor{gray}{66.7} & \textcolor{gray}{80.2} \\ 
$s2_{\text{B4}}(M, f^*)$~(rep.~\cite{kordopatis2021leveraging})~+~RRM   & \textcolor{gray}{18.6} &  \textcolor{gray}{41.8}    & \textcolor{gray}{55.3}     & \textcolor{gray}{69.2}     & \textcolor{gray}{82.7}     \\ 
MvMF$_{\text{B4}}$~(rep. \cite{kordopatis2021leveraging}) + RRM   & \textcolor{gray}{21.9} &  \textcolor{gray}{44.3}    & \textcolor{gray}{55.3}     & \textcolor{gray}{67.5}     & \textcolor{gray}{81.9}     \\ \hline
$s2_{\text{B4}}(M, f^*)$~(rep.)   & 14.3  & 42.6  &  55.7   & \textbf{71.3}  & 81.9   \\
$\text{\textbf{\acrshort{ah}}}_{\text{B4}}(\{100,125,250\}, f)$                                                                                                    &  16.9 & 42.6 & \textbf{56.1} & 69.6  &  \textbf{84.8} \\
$\text{\textbf{\acrshort{ah}}}_{\text{B4}}(\{50,75,100\}, f)$                                                                                                    &  19.4 & 41.2 & \textbf{56.1} & 68.0  & 81.0 \\ \hline
$s2(M, f)$   & 14.8 & 39.7 & 49.8 & 64.1 & 79.7 \\
$s2(M, f^*)$ (rep.)   & 15.2 & 40.9 &  51.9 & 65.8 & 80.6 \\ 
$\text{\textbf{\acrshort{ah}}}(\{100,125,250\}, f)$   &  15.2 & 36.7 & 48.1 & 64.1 &  78.1  \\
$\text{\textbf{\acrshort{ah}}}_a(\{100,150,250\}, f)$   & 12.7 & 37.1 & 48.1 & 64.6 &  78.1  \\
$\text{\textbf{\acrshort{ah}}}(\{50,75,100\}, f)$   & 16.0 & 38.4 & 49.4 & 63.7 &  78.5  \\
\bottomrule
\end{tabularx}
\end{center}
\end{table}

\paragraph{State-of-the-Art Partitioning:}
To evaluate the effectiveness of the proposed \acrshort{ah} to the commonly used \emph{s2} partitioning~--~which currently leads to state-of-the-art results~\cite{kordopatis2021leveraging}~--~we fix the entire setup and only compare the respective \emph{partitioning} methods.
As $s2(M, f^*)$~\cite{MuellerBudack2018GeolocationEstimationOfPhotosUsingAHierarchicalModelAndSceneClassification} provides state-of-the-art results without the usage of ensembles or other additional extensions, we reproduce the results using a \textit{ResNet} with 50 instead of 101 layers for a fair comparison.
The reproduced results~($s2(M, f^*)$~(rep.)) are slightly better than the original~(even using a less complex model) which is caused by the modified training procedure. 
The more complex \emph{EfficientNet} architecture improves results in general. However, it seems to have better capabilities for \acrshort{ah} to extract relevant features than with the $s2$ partitioning.
The advantage of \acrshort{ah} compared to \emph{s2} is only partially seen when using the \emph{ResNet} but tends to achieve slightly better or very comparable results otherwise.

For the model trained on cells with existing geodata~(area shape boundaries), the performance drops at finer scale but remains similar for the other scales since geodata is more often available in \acrshort{osm} for coarser regions.
While doubling the classes can improve the accuracy at street level~(less than 1\,km error) it leads to worse results on coarser scales, as also observed by previous work~\cite{Izbicki2019ExploitingTheEarthsSphericalGeometryToGeolocateImages}.

\textbf{State-of-the-Art Results:}
Especially when using the \emph{EfficientNet} architecture, \acrshort{ah} is superior to state-of-the-art models~\cite{Seo2018CPlaNetEnhancingImageGeolocalizationByCombinatorialPartitioningOfMaps, MuellerBudack2018GeolocationEstimationOfPhotosUsingAHierarchicalModelAndSceneClassification, Izbicki2019ExploitingTheEarthsSphericalGeometryToGeolocateImages} on almost all scales and test sets including re-implementations from \cite{kordopatis2021leveraging} that use the same underlying architecture and training dataset.
Superior results are achieved even without using ensembles and additional retrieval extensions~(colored gray) that could be considered for further improvements but is out of scope of this paper.

\begin{figure}[tbp]
\begin{center}
  \includegraphics[width=1.0\linewidth]{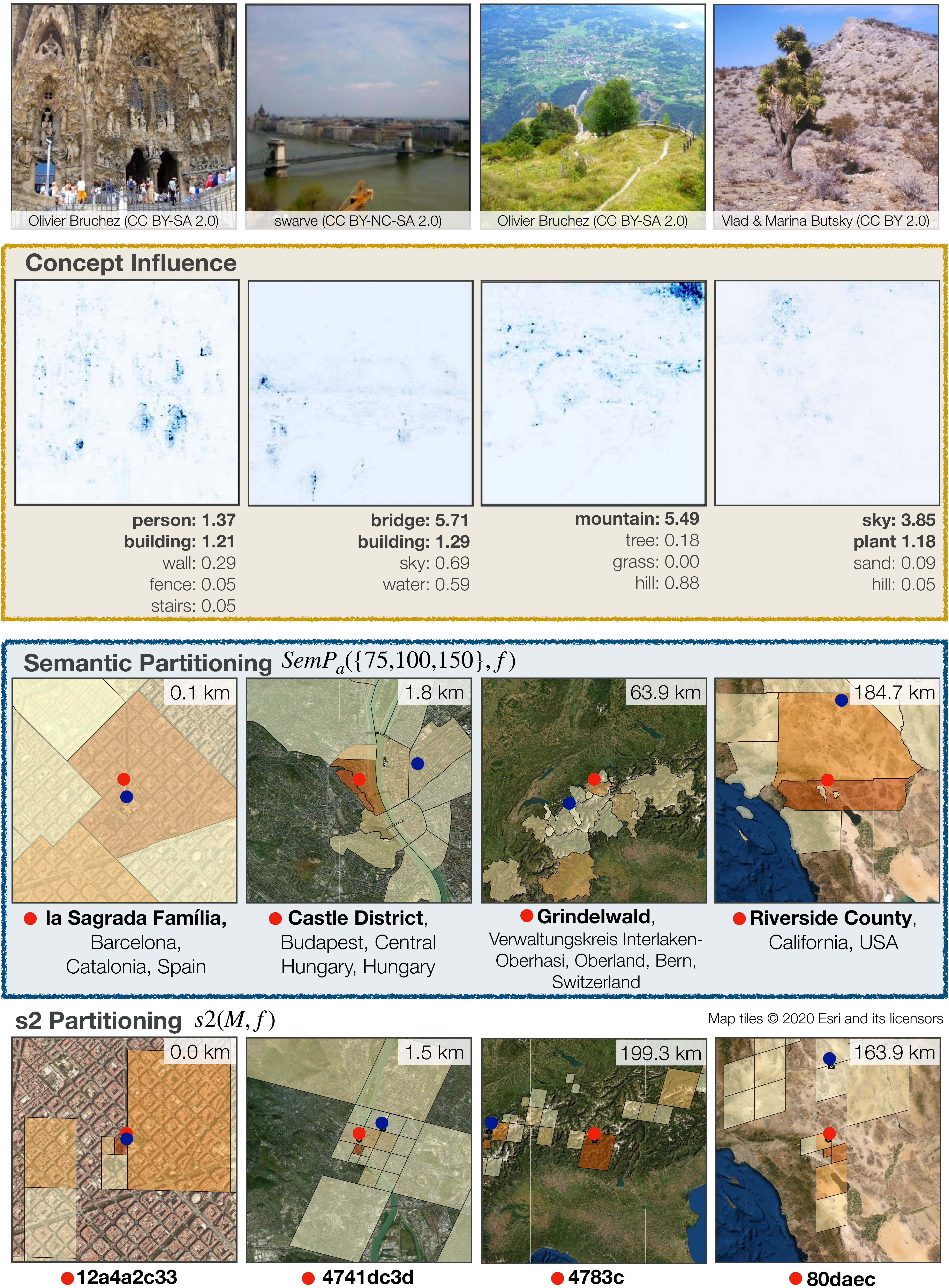}
\end{center}
\caption{Output of the proposed \emph{concept influence} metric in addition to the explanation map~\cite{Sundararajan2017AxiomaticAttributionForDeepNetworks} and qualitative comparison of predictions from the $\text{\acrshort{ah}}_a$ model and $s2(M, f)$~\cite{MuellerBudack2018GeolocationEstimationOfPhotosUsingAHierarchicalModelAndSceneClassification}~(last two rows). The top-25 classes and probabilities are visualized on the zoomed world map. 
Below is the predicted class label whereas the red marker is the predicted coordinate~(blue is ground-truth).
}
\label{fig:qualitative_ap_vs_s2}
\end{figure}

\textbf{Qualitative Comparison:}
Nevertheless, the goal is to develop a \emph{partitioning} that is intuitively comprehensible to humans, and yet delivers state-of-the-art results.
In the following, we discuss the findings from some qualitative results in detail with a focus on the interpretability of individual predictions.
In Figure~\ref{fig:qualitative_ap_vs_s2}~(the two lower rows) four examples from \emph{Im2GPS3k} are visualized where both $\text{\acrshort{ah}}_a(\{75,100,150\}, f)$ and $s2(M, f)$ share the same range of geolocational accuracy.
For each partitioning, the cells with the top probabilities~(max. 25) are colored in the zoomed region of the world map. 
The predicted label is depicted below the maps.
Both models are trained on two different types of partitionings and achieve similar geolocational accuracies. However, there are two main advantages of the proposed partitioning method over the \emph{s2} method during inference. 
First, not only a coordinate is provided but also the human-readable class label~(e.g., "\emph{la Sagrada Familia, Barcelona, Spain}") where its level of detail is ordered according to the semantic hierarchy and provided metadata.
Second, the visualization on the relevant part of the world map is much more clearly structured~(Figure~\ref{fig:qualitative_ap_vs_s2} third row) since the boundaries of the cells are not arbitrary selected but rather follow geographical borders which finally leads to a better understanding of a prediction. In line with this, the procedure of constructing smaller or more detailed cells is more natural in semantic partitioning \acrshort{ah}, since it follows a real hierarchical structure (e.g., from city to district), in contrast to the $s2$ algorithm with a hierarchy fixed to exactly four finer cells due to its underlying quad-tree.

\subsection{Understanding the Input Feature Importance}
\label{exp:ci}

\begin{table*}[bt!]
\caption{The \emph{concept influence}~(\emph{ci}$_{\text{median}}$) aggregated for each visual concept~($s$) in \acrshort{yfcc25600} and binned into spatial intervals depending on the achieved \acrshort{gcd} error in $km$. Presented are the highest and lowest $k=10$ concepts per spatial interval according to \acrshort{ci}.
}
\label{tab:ci_yfcc}
\setlength{\tabcolsep}{3.3pt}
\begin{center}
\small
\fontsize{8}{8}\selectfont

\begin{tabularx}{\linewidth}{lrr|lrr|lrr||lrr|lrr|lrr}
\toprule
\multicolumn{9}{c||}{top-10} & \multicolumn{9}{c}{lowest-10} \\
   \multicolumn{3}{c|}{$[0-25)$} & \multicolumn{3}{c|}{$[25-750)$} & \multicolumn{3}{c||}{$[750-2500)$} & \multicolumn{3}{c|}{$[0-25)$} & \multicolumn{3}{c|}{$[25-750)$} & \multicolumn{3}{c}{$[750-2500)$} \\ \hline
  $s$ &  $|s|$ & \emph{ci} &   $s$ &  $|s|$ & \emph{ci} &    $s$ &  $|s|$ & \emph{ci} &     $s$ &   $|s|$ & \emph{ci} &     $s$ &   $|s|$ & \emph{ci} &     $s$ &   $|s|$ & \emph{ci} \\
\midrule
      tower &    58 &   1.51 &  windowpane &   109 &   1.74 &  windowpane &   119 &   2.13 &     floor &  354 &   0.24 &     table &  119 &   0.15 &       base &   67 &   0.22 \\
        sky &  2114 &   1.29 &      animal &   142 &    1.4 &      animal &   162 &   1.38 &       car &  149 &   0.21 &     field &  150 &   0.15 &      chair &   70 &   0.19 \\
     animal &    58 &   1.26 &         sky &  2494 &   1.38 &         sky &  1518 &   1.15 &     earth &  486 &   0.16 &      path &   54 &   0.14 &       sand &   65 &   0.18 \\
   building &  1851 &   1.13 &       house &    92 &    1.3 &      person &  2218 &   1.15 &     water &  432 &   0.16 &     grass &  761 &   0.12 &      grass &  521 &   0.18 \\
   mountain &   439 &   1.09 &    mountain &   571 &   1.18 &    building &  1083 &    1.1 &     plant &  206 &   0.13 &   railing &   61 &   0.12 &      field &   78 &   0.17 \\
 windowpane &    51 &   1.04 &    airplane &    74 &   1.09 &    mountain &   279 &   1.09 &     grass &  380 &   0.11 &   bicycle &   58 &   0.11 &      table &  185 &   0.17 \\
     bridge &    86 &   0.96 &    building &  1658 &   1.06 &    airplane &    53 &   1.09 &      sand &   70 &   0.08 &     chair &   66 &   0.09 &    bicycle &   62 &   0.13 \\
     person &  1223 &   0.79 &      person &  2015 &   0.99 &      flower &   103 &   1.06 &     field &   66 &   0.08 &      road &  620 &   0.08 &       seat &   68 &   0.09 \\
 grandstand &    61 &   0.79 &      flower &   101 &   0.95 &        tree &  1247 &   0.97 &      road &  463 &   0.07 &      sand &  114 &   0.08 &   sidewalk &  198 &   0.09 \\
       wall &  1092 &   0.74 &        tree &  1889 &    0.9 &    painting &   141 &   0.91 &  sidewalk &  303 &   0.04 &  sidewalk &  294 &   0.07 &       road &  383 &   0.08 \\
\bottomrule
\end{tabularx}
\end{center}
\end{table*}

It is likely that certain visual concepts influence the predictions of a model differently at various geographical levels.
For instance, for landmarks, architectural features are probably dominant, whereas for landscapes with few visual clues to a concrete location, vegetation may provide some cues for a rough estimate. 
To investigate the proposed \acrshort{ci} score for several concepts on such geographic levels, we aggregate the \acrshort{ci} score for each concept~$s\in \mathbb{S}$.
We examine three geographic levels, where we assume the model predicts the location correctly based on different geographic properties.
In particular, we consider $[0-25)\,km$ for precisely predictable locations, $[25-750)\,km$ for regions, and $[750-2500)\,km$ for photos with few visual cues for a concrete location.
These are strict intervals where, for instance, a photo with a \acrshort{gcd} $<1\,km$ is not considered for the $[25-750)\,km$ interval.
To aggregate the \acrshort{ci}, we compute the median ~($ci_\text{median}$) and use it instead of the mean to ignore larger outlier(s). 
Please note, that similar conclusions can also be drawn from the mean value (see Appendix).
According to its definition~(Section~\ref{sec:ci}), the \emph{ci} score for concepts with geographic clues is expected to be greater than or equal to one, whereas the score for concepts without any hints should be close to zero.

\paragraph{Setup:}
We apply the \acrshort{ci} score to \acrshort{yfcc25600} 
due to its larger size compared to the test sets, and focus on the reproduced $s2(M, f^*)$ model in this experiment.
Since current segmentation models achieve high-quality results, we apply the \acrshort{hrnet}~\cite{Sun2019HighResolutionRepresentationsForLabelingPixelsAndRegions} which is trained on the \emph{ADE20k}~\cite{Zhou2017SceneParsingThroughAde20kDataset} dataset that contains 150 object classes (e.g., \emph{person, car, bottle}) and concepts for scene parsing (e.g., \emph{sky, ground, mountain}) and is therefore well suited. 
According to a study~\cite{Hooker2019ABenchmarkForInterpretabilityMethodsInDeepNeuralNetworks}, the method of \emph{Integrated Gradients}~\cite{Sundararajan2017AxiomaticAttributionForDeepNetworks} is chosen as explanation method, extended by \emph{SmoothGrad}~\cite{Smilkov2017SmoothGradRemovingNoiseByAddingNoise} which seeks to alleviate noise in explanation maps. 
Inserting random Gaussian noise in $n$ copies of the input image and then averaging the produced explanation maps cleans up artifacts. The noise parameter is set to $\sigma=0.15$ as suggested by the authors, but the sample size is reduced from $20$ to $5$ due to computing complexity without major result changes. 
Please note, that the variant as in~\cite{Hooker2019ABenchmarkForInterpretabilityMethodsInDeepNeuralNetworks} is used which squares each value before averaging.

\paragraph{Influence of Individual Concepts:}
\begin{figure}[tb!]
\begin{center}
  \includegraphics[width=1.0\linewidth]{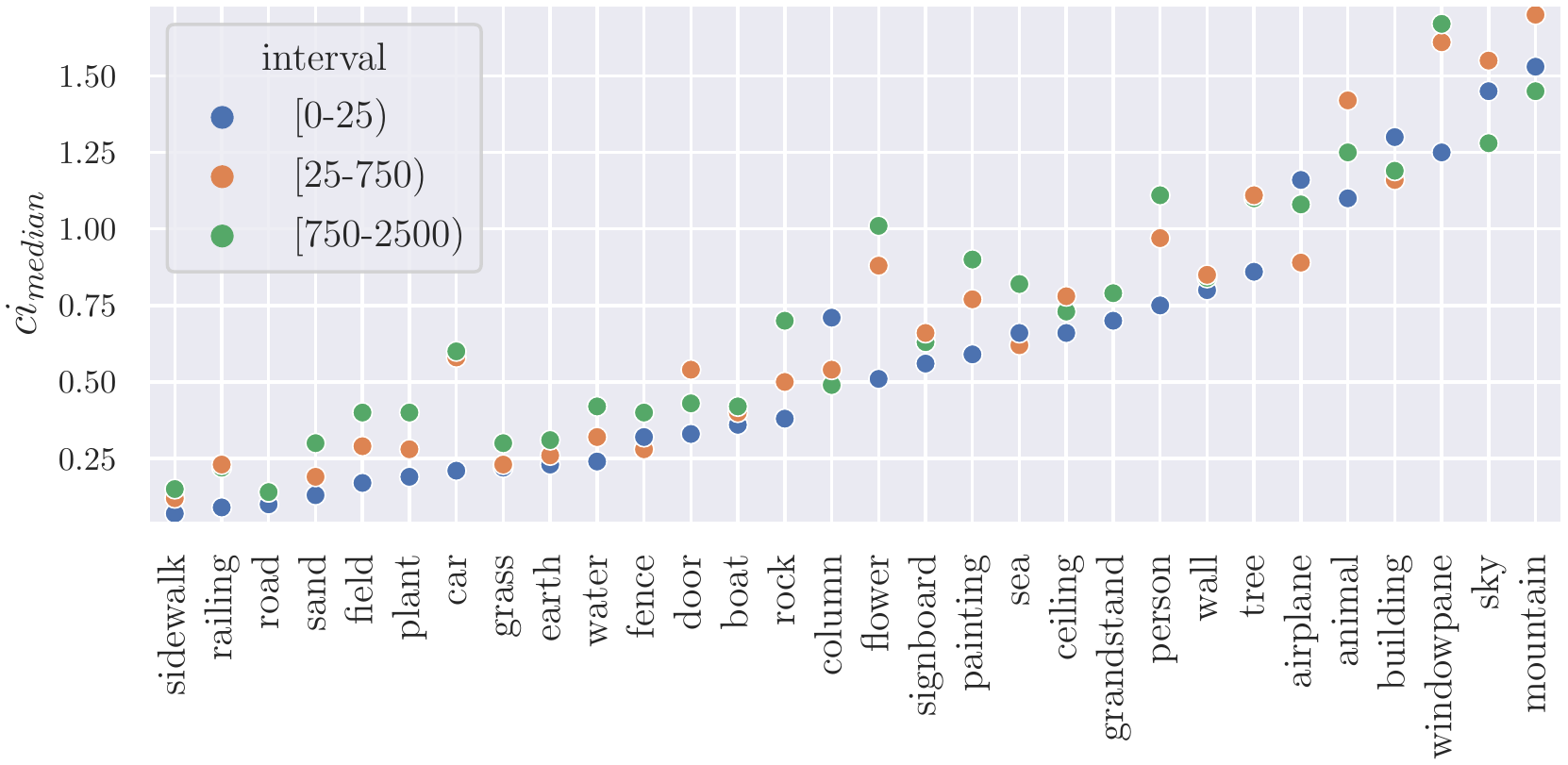}
\end{center}
  \caption{Absolute $ci_\text{median}$ scores for a selection of visual concepts divided in three geographic intervals.}
\label{fig:top_gap}
\end{figure}

We report results for the top-$k$ and lowest-$k$ \acrshort{ci} scores, i.e., concept label, $ci_\text{median}$, and the number of concepts~($|s|$) that fall into the evaluation interval.
Table~\ref{tab:ci_yfcc} shows results for concepts that occur in at least 50 images and where the morphological dilation is set to $\beta = 0$.
The complete table containing all concepts 
can be found in the Appendix.
Figure~\ref{fig:top_gap} shows the absolute $ci_\text{median}$ per concept for the respective geographic interval with a selection of concepts with high discrepancies.
The following observations can be made from Table~\ref{tab:ci_yfcc}.
Concepts like \textit{tower}, \textit{building}, \textit{bridge}, or \textit{mountain} have a high influence~($ci_\text{median} \gtrapprox 1$) at the $[0, 25)\,km$ interval and correspond to expected concepts to locate a place more precisely.
On the contrary, concepts like \textit{grass}, \textit{road}, \textit{water}, or \textit{car} have very limited influence~($ci_\text{median} \lessapprox 0.2$), which seems reasonable since these concepts are rather general concepts that are visually similar all over the world.
The concept of \emph{sky} has an initially surprisingly large influence on the prediction.
The two examples in Figure~\ref{fig:qualitative_ap_vs_s2}~(last two images) indicate that architectural details of buildings or peaks of mountain ranges can be relevant, i.e., the sky-touching concepts.
With the introduction of the morphological dilation~($m^\beta(\boldsymbol x, s)$) this area is covered. 
A repetition of this experiment with the enlarged area~($\beta=3$) confirmed this assumption. 
Since the \acrshort{ci} increases for \emph{windowpane}, \emph{person}, \emph{tree}, or \emph{animal} on higher geographical levels, such concepts are more relevant for rough estimations, where there are few visual clues to a more concrete location.
Lastly, the examples in Figure~\ref{fig:qualitative_ap_vs_s2} show an additional property of the presented metric for single instances. 
It does not determine the particular concept that is crucial for a prediction but rather which concepts are influential.

\section{Conclusions}
\label{sec:conclusion}
In this paper, we have presented a novel semantic photo geolocalization system that allows for the interpretation of results. 
To achieve this, we have proposed a semantic partitioning method that leads to an improved comprehensibility of predictions while at the same time achieving state-of-the-art results on common benchmark test sets.
In addition, we have suggested a novel metric to assess the importance of semantic visual concepts for a certain prediction to provide additional explanatory information, and to allow for a large-scale analysis of already trained models.

In the future, we plan to incorporate visual similarities between classes based on geographical features during optimization, e.g., derived from a knowledge base, since currently visual and spatial proximate classes are equally penalized as visual and spatial dissimilar classes.

\section*{Acknowledgement}
This project has partially received funding from the German Research Foundation (DFG: Deutsche Forschungsgemeinschaft, project number: 442397862).

\appendix

\section{Influence of Individual Concepts}\label{apdx:tab}

\subsection{Morphological Dilation $\beta$}
As mentioned in the paper, the concept \emph{sky} has an initially surprisingly large influence on the prediction.
We argued that sky-touching concepts could be responsible for this as the boundaries between concepts are likely of interest for the model.
To cover this area with the \acrshort{ci}, we have introduced the morphological dilation~($m^\beta$(x, s). 
We compute the \acrshort{ci}$_\text{median}$ both for $\beta=3\,px$ and $\beta=0\,px$ and those concepts, as $3\,px$ are sufficient to cover these boundaries.
In Table~\ref{tab:appendix_beta} the absolute difference between \acrshort{ci}$_\text{median}^{\beta=3}$ and \acrshort{ci}$_\text{median}^{\beta=0}$ is reported.
The significant increase for \emph{mountain} and \emph{tower} confirms our explanation. However, the same effect cannot be observed for \emph{building}.
The reason is that this concept covers a wide range of sub-concepts that are not investigated quantitatively in this work.
However, we are confident that the same effect can be observed when sub-concepts like skyscraper or stadium are investigated individually as qualitative examples indicate this.
With the release of source code we hope to foster further research in this area as we were only able to provide initial insights with the \acrshort{ci} in this paper.

\begin{table}[tbh!]
\caption{The absolute difference $\Delta$ of the \emph{concept influence}~(\emph{ci}$_{\text{median}}$) for morphological dilation ($\beta=3$ and $\beta=0$) aggregated for relevant sky-touching concepts in \acrshort{yfcc25600} and binned into spatial intervals depending on the achieved \acrshort{gcd} error in $km$. 
}
\label{tab:appendix_beta}
\begin{center}
\fontsize{8}{8}\selectfont
\begin{tabular}{lrrr}
\toprule
    concept &  $\Delta_\beta$ $[0-25)$ &  $\Delta_\beta$ $[25-750)$ &  $\Delta_\beta$ $[750-2500)$ \\
\midrule
      sky &         +0.16 &           +0.17 &             +0.13 \\
 mountain &         +0.44 &           +0.52 &             +0.36 \\
 building &         +0.17 &           +0.10 &             +0.09 \\
    tower &         +0.34 &           -     &             -     \\
\bottomrule
\end{tabular}
\end{center}
\end{table}

\subsection{Aggregation Operator}
To aggregate the \acrshort{ci} score over all images for a certain concept, the \emph{median} or \emph{average} operator was suggested.
In Table~\ref{tab:admin_test_all_appendix_1},\ref{tab:admin_test_all_appendix_2} both variants are reported for all concepts where at least ten images are available whereas only the top/last-k concepts were reported in the main paper with at least 50 images.

\begin{table}[t!]
\caption{The \emph{concept influence} aggregated for each visual concept $s$ (start letter a-m) in \emph{YFCC-Val26k} and binned into spatial intervals depending on the achieved \emph{GCD} error in $km$. Presented are concepts of the dataset where at least 10 images are available.}
\label{tab:admin_test_all_appendix_1}
\setlength{\tabcolsep}{2pt}
\setlength\extrarowheight{-3pt}
\begin{center}
\small
\fontsize{7}{10}\selectfont

\begin{tabular}{lrrrrrr}
\toprule
{} & \multicolumn{3}{r}{median} & \multicolumn{3}{r}{mean} \\
{} & [0-25) & [25-750) & [750-2500) & [0-25) & [25-750) & [750-2500) \\
\midrule
airplane   &   1.32 &     1.09 &       1.09 &   1.92 &     1.53 &       2.11 \\
animal     &   1.26 &     1.40 &       1.38 &   2.20 &     2.11 &       2.02 \\
apparel    &        &     0.15 &       0.12 &        &     0.29 &       0.23 \\
armchair   &        &          &       0.22 &        &          &       0.34 \\
awning     &   0.18 &     0.35 &       0.54 &   0.51 &     1.00 &       1.23 \\
bag        &        &          &       0.20 &        &          &       0.67 \\
ball       &        &     0.59 &       0.70 &        &     0.89 &       0.70 \\
bannister  &   0.13 &     0.21 &            &   0.72 &     0.60 &            \\
bar        &        &          &       0.51 &        &          &       0.85 \\
barrel     &        &     0.18 &       0.37 &        &     0.74 &       0.61 \\
base       &   0.37 &     0.40 &       0.22 &   0.70 &     0.78 &       0.68 \\
bed        &        &     0.19 &       0.67 &        &     0.41 &       0.65 \\
bicycle    &   0.11 &     0.11 &       0.13 &   0.41 &     0.54 &       0.57 \\
boat       &   0.30 &     0.30 &       0.28 &   0.75 &     0.77 &       1.01 \\
book       &        &          &       0.31 &        &          &       0.98 \\
bottle     &        &     0.19 &       0.76 &        &     0.70 &       0.93 \\
box        &   0.48 &     0.30 &       0.57 &   0.88 &     0.51 &       0.98 \\
bridge     &   0.96 &     0.29 &       0.83 &   1.52 &     0.91 &       1.23 \\
building   &   1.13 &     1.06 &       1.10 &   1.57 &     1.49 &       1.56 \\
bulletin   &   0.73 &     0.33 &       0.44 &   0.89 &     0.96 &       0.49 \\
bus        &   0.69 &     0.82 &       0.89 &   0.86 &     1.25 &       0.75 \\
cabinet    &        &     0.28 &       0.49 &        &     0.70 &       1.21 \\
car        &   0.21 &     0.54 &       0.52 &   0.79 &     0.97 &       0.89 \\
case       &        &          &       0.18 &        &          &       0.55 \\
ceiling    &   0.61 &     0.68 &       0.59 &   1.10 &     1.21 &       0.95 \\
chair      &   0.09 &     0.09 &       0.19 &   0.79 &     0.46 &       0.52 \\
clock      &        &     0.35 &       0.89 &        &     0.53 &       1.30 \\
column     &   0.71 &     0.45 &       0.25 &   0.91 &     1.26 &       0.83 \\
computer   &        &          &       0.29 &        &          &       0.68 \\
counter    &        &          &       0.29 &        &          &       0.95 \\
crt        &   0.85 &     1.40 &       0.48 &   1.42 &     1.13 &       0.94 \\
curtain    &   0.25 &     0.37 &       0.30 &   0.52 &     0.84 &       0.80 \\
desk       &        &     0.53 &       0.21 &        &     0.75 &       0.63 \\
dirt       &   0.04 &     0.03 &       0.03 &   0.25 &     0.10 &       0.11 \\
door       &   0.27 &     0.47 &       0.30 &   0.85 &     0.77 &       0.77 \\
earth      &   0.16 &     0.17 &       0.22 &   0.47 &     0.50 &       0.55 \\
fence      &   0.24 &     0.17 &       0.32 &   0.73 &     0.62 &       0.72 \\
field      &   0.08 &     0.15 &       0.17 &   0.35 &     0.49 &       0.64 \\
floor      &   0.24 &     0.20 &       0.29 &   0.67 &     0.54 &       0.61 \\
flower     &   0.46 &     0.95 &       1.06 &   1.14 &     1.39 &       1.22 \\
food       &   0.93 &     0.66 &       0.62 &   1.30 &     1.00 &       0.84 \\
fountain   &   0.39 &     0.78 &       0.25 &   1.06 &     1.29 &       0.82 \\
grandstand &   0.79 &     1.02 &       0.86 &   1.01 &     1.22 &       1.09 \\
grass      &   0.11 &     0.12 &       0.18 &   0.41 &     0.44 &       0.52 \\
hill       &   0.47 &     0.44 &       0.68 &   1.05 &     1.05 &       1.05 \\
house      &   1.98 &     1.30 &       1.29 &   1.92 &     1.85 &       1.48 \\
hovel      &        &     0.77 &       0.63 &        &     1.30 &       1.39 \\
land       &        &     0.77 &            &        &     1.67 &            \\
minibike   &   0.36 &     0.63 &       0.15 &   0.91 &     0.95 &       0.52 \\
mountain   &   1.09 &     1.18 &       1.09 &   1.72 &     1.82 &       1.67 \\
\bottomrule
\end{tabular}

\end{center}
\end{table}

\begin{table}[t!]
\caption{The \emph{concept influence} aggregated for each visual concept $s$ (start letter n-z) in \emph{YFCC-Val26k} and binned into spatial intervals depending on the achieved \emph{GCD} error in $km$. Presented are concepts of the dataset where at least 10 images are available.}
\label{tab:admin_test_all_appendix_2}
\setlength{\tabcolsep}{2pt}
\setlength\extrarowheight{-3pt}
\begin{center}
\small
\fontsize{7}{10}\selectfont

\begin{tabular}{lrrrrrr}
\toprule
{} & \multicolumn{3}{r}{median} & \multicolumn{3}{r}{mean} \\
{} & [0-25) & [25-750) & [750-2500) & [0-25) & [25-750) & [750-2500) \\
\midrule
painting   &   0.52 &     0.78 &       0.91 &   0.82 &     0.96 &       1.14 \\
palm       &   0.67 &     0.72 &       1.12 &   1.09 &     1.84 &       1.69 \\
path       &   0.10 &     0.14 &       0.09 &   0.62 &     0.87 &       0.76 \\
person     &   0.79 &     0.99 &       1.15 &   1.17 &     1.33 &       1.45 \\
pier       &   0.09 &     0.05 &            &   0.44 &     0.42 &            \\
plant      &   0.13 &     0.24 &       0.38 &   0.49 &     0.57 &       0.66 \\
plate      &        &     0.83 &       0.54 &        &     1.30 &       0.81 \\
plaything  &        &     0.55 &       0.45 &        &     1.50 &       1.18 \\
pole       &   0.02 &     0.19 &            &   0.31 &     0.46 &            \\
poster     &        &     0.58 &       0.47 &        &     1.33 &       1.53 \\
railing    &   0.11 &     0.12 &       0.13 &   0.42 &     0.53 &       0.47 \\
river      &   0.10 &     0.07 &       0.10 &   0.49 &     0.33 &       0.54 \\
road       &   0.07 &     0.08 &       0.08 &   0.30 &     0.40 &       0.37 \\
rock       &   0.37 &     0.49 &       0.67 &   0.86 &     1.05 &       1.24 \\
runway     &   0.15 &     0.28 &       0.15 &   0.37 &     0.54 &       0.38 \\
sand       &   0.08 &     0.08 &       0.18 &   0.36 &     0.45 &       0.67 \\
sculpture  &   0.58 &     0.68 &       0.58 &   1.44 &     0.92 &       1.14 \\
sea        &   0.40 &     0.32 &       0.49 &   0.86 &     0.84 &       0.91 \\
seat       &   0.18 &     0.25 &       0.09 &   0.50 &     0.50 &       0.35 \\
shelf      &        &     0.43 &       0.69 &        &     1.00 &       0.87 \\
ship       &   0.50 &     0.81 &       0.60 &   1.12 &     1.03 &       0.93 \\
sidewalk   &   0.04 &     0.07 &       0.09 &   0.31 &     0.51 &       0.40 \\
signboard  &   0.60 &     0.64 &       0.57 &   1.27 &     1.31 &       1.26 \\
sink       &        &     1.05 &       0.97 &        &     1.09 &       1.04 \\
sky        &   1.29 &     1.38 &       1.15 &   1.96 &     2.08 &       1.78 \\
skyscraper &   0.30 &          &            &   0.72 &          &            \\
sofa       &        &          &       0.22 &        &          &       0.43 \\
stage      &   0.62 &     0.31 &       1.19 &   1.15 &     0.55 &       1.41 \\
stairs     &   0.03 &     0.05 &       0.21 &   0.21 &     0.31 &       0.51 \\
stairway   &   0.13 &     0.13 &       0.17 &   0.41 &     0.64 &       0.39 \\
table      &   0.17 &     0.15 &       0.17 &   0.67 &     0.53 &       0.63 \\
tank       &        &     0.41 &       0.48 &        &     0.56 &       0.72 \\
tent       &   1.13 &     0.25 &       0.40 &   1.51 &     0.78 &       0.84 \\
tower      &   1.51 &     0.73 &       1.58 &   2.21 &     1.15 &       2.27 \\
tray       &        &     0.61 &       0.55 &        &     0.90 &       0.95 \\
tree       &   0.71 &     0.90 &       0.97 &   1.25 &     1.43 &       1.53 \\
truck      &   0.75 &     0.65 &       0.92 &   0.94 &     1.04 &       1.13 \\
van        &        &          &       0.49 &        &          &       0.82 \\
wall       &   0.74 &     0.78 &       0.78 &   0.92 &     0.98 &       0.98 \\
water      &   0.16 &     0.21 &       0.33 &   0.48 &     0.57 &       0.58 \\
waterfall  &   0.87 &     0.38 &       0.50 &   2.18 &     0.99 &       0.93 \\
windowpane &   1.04 &     1.74 &       2.13 &   2.28 &     2.63 &       2.75 \\
\bottomrule
\end{tabular}

\end{center}
\end{table}

\section{Extended MP-16 Dataset (EMP-16)}\label{sec:dataset}

Assigning a single GPS coordinate to each type of image is debatable in terms of content, but corresponds to the traditional task where one image is assigned to exactly one coordinate.
At present, even specialized models struggle because of the complexity of the task, making a discretization of coordinates with semantic context necessary in our opinion.

As formally described in the main paper
, a reverse geocoder can be used to assign each coordinate an address, which is our required information to construct the \acrfull{ah}.
Thanks to \acrshort{osm} many additional information~(e.g. \emph{geometry}) are available for each location of an address.
To foster future research in this context, we extend the \acrfull{mp16}~\cite{Larson2017TheBenchmarkingInitiativeForMultimediaEvaluationMediaEval2016} dataset for photo geolocalization~(image-coordinate pairs) with those additional information and call it \emph{Extended MP-16} (\emph{EMP-16}).
This section provides a short insight of the construction and resulting dataset.

As described in the paper, we utilize the \acrshort{mp16}~\cite{Larson2017TheBenchmarkingInitiativeForMultimediaEvaluationMediaEval2016} dataset that is a subset from the \acrfull{yfcc100m}~\cite{Thomee2016YFCC100MTheNewDataInMultimediaResearch}. 
Its only restriction is that an image contains a \acrshort{gps} coordinate, thus it contains images of landmarks, landscape images, but also images with little to no geographical cues. Like~\citet{Vo2017RevisitingIm2gpsInTheDeepLearningEra} images are excluded from the same authors as in the \emph{Im2GPS3k} test set and duplicates are removed resulting in a dataset size of 4,723,695 image-coordinate pairs.

\begin{table*}[bt!]
\caption{
Example for additional meta-information available for extracted locations from \acrshort{osm}: Given the GPS coordinate of an image-coordinate pair, each location of the resulting address hierarchy~(e.g.~[\emph{W438331516}, \emph{W13470104}, ...]) contains additional context information. \emph{Geometry} contains single points, multi-lines, polygons and multi-polygons.}
\label{tab:dataset_extract}
\small
\begin{center}
\setlength\extrarowheight{-3pt}
\setlength{\tabcolsep}{1pt}
\begin{tabularx}{\linewidth}{X|llllll}
\toprule
id &                                         W438331516 &                                          W13470104 &                                            R112100 &                                            R396479 &                                            R165475 &                                            R148838 \\
\midrule
localname   &                                     The Queen Mary &                                        Windsor Way &                                         Long Beach &                                 Los Angeles County &                                         California &                                      United States \\
category    &                                           historic &                                            highway &                                           boundary &                                           boundary &                                           boundary &                                           boundary \\
type        &                                               ship &                                            service &                                     administrative &                                     administrative &                                     administrative &                                     administrative \\
admin\_level &                                                 15 &                                                 15 &                                                  8 &                                                  6 &                                                  4 &                                                  2 \\
isarea      &                                               True &                                              False &                                               True &                                               True &                                               True &                                               True \\
wikidata    &                                            Q752939 &                                               NA &                                             Q16739 &                                            Q104994 &                                                Q99 &                                                Q30 \\
wikipedia   &                                               NA &                                               NA &                          en:Long Beach, California &                  en:Los Angeles County, California &                                      en:California &                                   en:United States \\
population  &                                               NA &                                               NA &                                             469450 &                                               NA &                                           37253956 &                                          324720797 \\
place       &                                               NA &                                               NA &                                               city &                                             county &                                              state &                                            country \\
geometry    &  [[[-118.191... &  [[-118.1... &   [[[-118.248... &   [[[[-1... &  [[[[-1... &  [[[[-1... \\
\bottomrule
\end{tabularx}
\end{center}
\end{table*}

For reverse geocoding we have applied \emph{Nominatim}~v3.4\footnote{\url{https://github.com/osm-search/Nominatim}} on a planet-scale \acrshort{osm} installation with a dump from 22nd May 2020.
After initial reverse geocoding and ignoring postal codes, the resulting multi-label dataset contains 1,191,616 unique locations which is not useful for this dataset size. 
To reduce the number, an initial filtering step with $\tau_\mathbb{D}=50$ is applied that removes non-frequent locations resulting in manageable 46,240 unique locations.
Each unique location which is identified by \emph{osm\_type} amd \emph{osm\_id} has a set of additional attributes such as \emph{localname}, \emph{geometry} or sometimes even links to the \emph{wikidata}~\cite{vrandevcic2014wikidata} knowledge base.
The extended information for one sample (image-coordinate pair) of the \acrshort{mp16} dataset is presented in Table~\ref{tab:dataset_extract}.
Since only the \emph{id}s, \emph{localname}s and available \emph{geometry} are used in this work, a full description and statistics of available attributes goes beyond of the scope of this paper.

\section{Model Pre-Training}\label{apdx:pretraining}
It was mentioned that the initial weights from all trained models are not selected randomly or taken from a model that was trained on \textit{ImageNet}~\cite{Russakovsky2015ImagenetLargeScaleVisualRecognitionChallenge} as commonly chosen, but with weights of a pre-trained model which already learned geo-related features to reduce the training time.
This sections provides missing details about the weight initialization.

\paragraph{Task and Data:} 
To construct a simple geolocation classification problem, each image is associated with the country~(including maritime borders) the image was taken. 
Note, that an image is associated to its country by checking whether its respective coordinate is within the area that the country covers and no function of the method presented in Section~3 like Reverse Geocoding is applied.
As data basis serves the MP-16~\cite{Larson2017TheBenchmarkingInitiativeForMultimediaEvaluationMediaEval2016} dataset for training and \acrshort{yfcc25600} for validation. 
Due to class imbalance, the USA is divided into individual states and a country is only considered if there are at least 1,000 images for training which results in 173 classes in total.

\paragraph{Network Training and Results:} One ResNet-50 is initialized with \emph{ImageNet} weights and optimized using the \acrshort{sgdr}~\cite{Loshchilov2017SGDRStochasticGradientDescentWithWarmRestarts} method with weight decay of $10^{-4}$, a momentum of $0.9$, a maximum learning rate of $5 \times 10^{-3}$ and one cycle is performed three times per epoch. The model is trained for ten epochs and is validated every 4,000 steps for a batch size of 128 (512,000 images).
The same image pre-processing and data augmentation methods are applied as described in the paper and below.
The standard classification loss~(cross-entropy) is used as loss function, i.e. one-hot-encoded vectors as ground-truth class) and the best model on the validation set is selected.
After training, this trained model achieves a top-1 accuracy of 0.24 percent (0.49 top-5).

\section{Image Augmentation}
During training, images are augmented as follows to prevent overfitting: Images are initially resized to a minimum edge size of 320 pixels. After that, we crop a region of random size (between $0.66$ and $1.0$ of original size) and random aspect ratio (3/4 to 4/3) of the original aspect ratio. This crop is finally resized to $224 \times 224$ pixels~($300 \times 300$ for \emph{EfficientNet}) and randomly flipped horizontally. For validation, the image is resized to a minimum edge size of $256$~($320$) pixels and then a center crop of size $224$~($300$) pixels is made.


{\small
\setlength{\bibsep}{0pt} 
\bibliographystyle{ieee_fullname} 
\bibliography{main}
}

\end{document}